\documentclass[conference]{IEEEtran}
\IEEEoverridecommandlockouts
\usepackage{cite}
\usepackage{amsmath,amssymb,amsfonts}
\usepackage{algorithmic}
\usepackage{graphicx}
\usepackage{textcomp}
\usepackage{xcolor}
\usepackage{hyperref}  
\usepackage{array} %for fixed table column length
\usepackage{titlesec}%for additional sub-section parts
\usepackage{afterpage}
\usepackage{lineno}%line numbers
\usepackage{array} %for fixed table column length
\usepackage{adjustbox}
\usepackage{multirow} %for multirow tables
\usepackage{bigstrut} %for multirow tables 
\def\BibTeX{{\rm B\kern-.05em{\sc i\kern-.025em b}\kern-.08em
    T\kern-.1667em\lower.7ex\hbox{E}\kern-.125emX}}
\begin{document}

\title{Pilot Study on Student Public Opinion Regarding GAI}

\author{\IEEEauthorblockN{ William Franz Lamberti \IEEEauthorrefmark{1}, Sunbin Kim\IEEEauthorrefmark{2}
Samantha Rose Lawrence\IEEEauthorrefmark{3}}
\IEEEauthorblockA{
\textit{Computational and Data Sciences}\\
\textit{College of Science}\\
\textit{George Mason University }\\
Fairfax, VA, United States \\
Email: \IEEEauthorrefmark{1}wlamber2@gmu.edu,
\IEEEauthorrefmark{2}skim253@gmu.edu
\IEEEauthorrefmark{3}slawre2@gmu.edu,
}}

\maketitle
\IEEEpubidadjcol

\begin{abstract}
The emergence of generative AI (GAI) has sparked diverse opinions regarding its appropriate use across various domains, including education. This pilot study investigates university students' perceptions of GAI in higher education classrooms, aiming to lay the groundwork for understanding these attitudes. With a participation rate of approximately 4.4\%, the study highlights the challenges of engaging students in GAI-related research and underscores the need for larger sample sizes in future studies. By gaining insights into student perspectives, instructors can better prepare to integrate discussions of GAI into their classrooms, fostering informed and critical engagement with this transformative technology. 
\end{abstract}

\begin{IEEEkeywords}
Artificial Intelligence, Generative, Education, Public Opinion%, aim for 6k-10k words
\end{IEEEkeywords}

\section{Introduction}

In recent years, the intersection of art and technology has witnessed a significant transformation, marked by the emergence of generative AI (GAI) as a powerful creative tool. By 2025, a notable trend emerged where people began using OpenAI's Chat GPT to transform their photographs into Studio Ghibli-style artworks, sharing these creations widely on social media platforms \cite{kircher_people_2025}. This trend not only highlighted the artistic capabilities of AI but also sparked discussions about the influence of GAI on traditional art forms and the authority of art styles.  Despite the creative potential of AI-generated art, it has raised concerns about creative labor, transparency, and copyright infringement \cite{jeegandhi2025topic}. This pilot study aims to explore the perspectives of university students enrolled in introductory computing courses regarding GAI, providing insights into their perceptions and informing future course designs.  We also report the participation rates to help inform future work and studies. By examining these attitudes, educators can better understand how to present GAI to maximize their communication of related topics.

In 2025, one of the major trends that users ask Open AI's Chat GPT is to convert their photographs into Studio Ghibli style\cite{kircher_people_2025}. Considering the historical context, since the 1960s, art generated by computer programs has been around, when British painter Harold Cohen created the AARON program to generate art based upon information it was fed\cite{cengel_2024_smithsonian}. In 2014, with the introduction of Generative Adversarial Networks, interest in GAI art renewed\cite{goodfellow_generative_2014}.  AI usage in generating artworks based upon the paintings of influential historical artists has been coined "Zombie Art"\cite{Hassine_Neeman_2019}. In 2022, the judges at the Colorado State Fair unknowingly awarded first place to an AI generated image, sparking national debate around the use of AI generated art\cite{nytroose2022}.

However, the main concerns regarding AI generated art span creative labor, transparency, and ownership \cite{jeegandhi2025topic} including copyright infringement. In 2023, The New York Times filed a lawsuit against OpenAI and Microsoft concerning training the chatbots with its articles might cause hallucinations even without receiving their permission \cite{grynbaum_mac_2023aicopyright}. Similarly, Disney and Universal had sued Midjourney, a GAI model that produces images and videos based from textual prompts \cite{midjourney2024guide}, for copyright infringement in 2025\cite{barnes2025disney}. On the other hand, it has been argued that GAI works can meet the philosophical definitions of art, even amid unease around their legitimacy\cite{ivanova2025ai}.

Previous studies have examined public perceptions of AI.  Notably, a recent report from the Pew Research Center observed distinct differences in attitudes toward AI across age groups, particularly U.S. adults under 30 engage with AI more frequently than adults aged 65 and above \cite{fuentes_how_2025}.  About 32,707 comments from 44 YouTube videos were analyzed investigating public expressions of concern about job displacement alongside expressing expectations for generative AI's potential to assist solving complex problems \cite{mahmoud2025}.  Controversies also exist within educational contexts. A student at Northeastern University requested a tuition refund in 2025 having discovered that her professor had used AI to develop the course materials, which even included the prompts the professor used in conversations with the AI \cite{hill_professors_2025}.

However, relatively less research has been conducted in traditional classrooms, including studies that examine students' public opinions about GAI -- from general perceptions of AI to its use within their courses. Therefore, this research is significant as the study specifically investigates undergraduate students' attitudes primarily composed of a demographic that is more familiar with the use of GAI \cite{fuentes_how_2025}.  Thus, obtaining enough data from this cohort is paramount.

There is substantial work regarding various topics within the educational framework and student response rates \cite{fosnacht_how_2017, wu_response_2022, sax_assessing_2003}.  However, these studies seem to be primarily concerned with larger surveys and not student participation in individual classrooms itself.  Other studies on GAI in the classroom have reported participation counts\cite{wang_learning_2025}, but it does not state how many did not participate.  The one study that we are aware of regarding GAI in the classroom participation rates reported a participation rate of about 28\% \cite{lamberti_pilot_2025}.  However, when we utilize only those classes that are the most similar, we obtain a participation rate of about 22\%.  

The purpose of this pilot study is to research the perspectives of university students who are taking introductory course for computing offered by the Department of Computational and Data Sciences (CDS) at George Mason University (GMU) regarding GAI in diverse contexts. By understanding the complexities of student opinion on GAI, instructors will be able to gain a clearer understanding of students' current perceptions of AI use and incorporate these insights into the design of future courses.  Further, the paper also reports and discusses the participation rate of students in the study.  This will help to inform what class sizes are needed to collect enough data.

\section{Methods}

\subsection{Context For AI}

It is important to define some terminology and provide definitions.  These definitions will help to characterize and better understand what items we are and are not mentioning.  The two major items are: AI and GAI.  While there does not appear to be a unified consensus on these definitions\cite{thierer_artificial_2017}.  We recognize that some might have differing definitions, but we will utilize the following definitions and descriptions defined previously\cite{lamberti_artificial_2024}. 

\subsection{Survey Design}

Both sections were given the same setup where they were asked the 7 questions, watched a video, and then asked the same 7 questions again.  We then compared before and after the video to observe if opinions changed after watching the video.

\subsection{Selection of Participants}

Participants were from CDS classes offered at GMU during the Fall 2025 session.  These courses are GMU Common Core Courses.

Student participation in the study was not required for taking the class.  Those that wished to not participate in the class were not treated differently nor removed from the class.  Additional details on the IRB details are provided in the Appendix.  

The following tables provides some high level information regarding the composition of students that agreed to participate in the study.

\begin{table}
    \centering
    \begin{tabular}{c|cc}
       Course & Modality  & Details \\\hline
       130 DL3  & Async &  2 classes for 1.5 hours per week\\
       130 DL4 & Async &  1 class for 3 hours per week
    \end{tabular}
    \caption{Information regarding the CDS 130 courses that were selected for the study.}
    \label{tab:my_label}
\end{table}

\subsection{Statistical Methods}

Part of our investigation required a deeper understanding of the response rate for participation in our study.  Since participants can either participate or not, this setup follows a Binomial distribution.  The Binomial distribution is well studied, but is provided here for completeness:

\begin{equation}\label{eq: binom}
    f(X=x) = {n \choose x}p^x(1-p)^{n-x}
\end{equation}

\noindent for $x = 0, 1, ..., n$, where $n$ is the number of trials\cite{bhattacharyya_statistical_1977}.  In our case, $p = $ the proportion of student participants who agree to share their data for our research.  To understand what $p$'s value is, we will construct 95\% confidence intervals (CI).  The manner of calculating these CIs has been debated for some time\cite{clopper_use_1934, newcombe_two-sided_1998, thulin_cost_2014, johnson_univariate_2005}.  We will be utilizing the Clopper-Pearson CIs (which is usually referred to as the `exact' CIs) since they are considered one of the most accurate and conservative methods\cite{newcombe_two-sided_1998}.  The lower and upper bounds of a CI at a given significance level ($\alpha$) are found by solving, respectively:

\begin{equation}\label{eq: ci_lower}
    \sum_{X=x}^n {n \choose x}p_l^x(1-p_l)^{n-x} = \frac{\alpha}{2},
\end{equation}

\begin{equation}\label{eq: ci_upper}
    \sum_{X=0}^x {n \choose x}p_u^x(1-p_u)^{n-x} = \frac{\alpha}{2}. 
\end{equation}

\noindent  We will simply refer to the Clopper-Pearson or exact CI as CI moving forward. 

We calculated these using R's \texttt{binom.test()}.  Comparing CIs is performed in the usual manner.  When comparing two CIs and they do not overlap, there is evidence that the $p$'s of the two CIs differ.  If they do overlap, there is no evidence that the $p$'s of the CIs differ. The response rate of this study will be compared to the relevant response rate of a similar, but different, pilot study performed in Summer 2025 related to GAI in the classroom\cite{lamberti_pilot_2025}. 

\section{Results}

We first discuss the response rate.  After we go through the high-level takeaways from this survey.  We then analyze each question sequentially. 

\subsection{Response Rate}

\begin{table}[ht!]
\centering
\begin{tabular}{|l|l|l|l|}
\hline
\textbf{Course \#} & \textbf{Section} & \textbf{\# of Enrolled} & \textbf{\# of Participating}  \\
\hline
CDS 130 & DL3 & 35 & 1 \\
CDS 130 & DL4 & 33 & 2 \\ \\
        & Totals & 68 & 3 \\
\hline
\end{tabular}
\caption{Participation and enrollment counts for the selected courses.  The number of enrolled students data was obtained using public data via \url{https://ssbstureg.gmu.edu/StudentRegistrationSsb/ssb/term/termSelection?mode=search}. }
\label{tab:course_enrollment}
\end{table}

Table \ref{tab:course_enrollment}'s response counts correspond to an overall response rate of about 0.044.  The associated exact 95\% CI for this rate is about (0.009, 0.124). The 95\% CI for a Summer 2025 GAI study was about (0.142, 0.429).  Since the CIs do not overlap, we have evidence that the participation rates differ.  

\subsection{High-Level Insights}

Since the response counts are relatively low, the insights gleaned from the questions is severely weakened.  We will still provide a na\"ive analysis, but we do not recommend putting a strong weight on these insights at this time.  2 of the 6 questions retained for this analysis had an observed change in the distributions of responses. Question 5 is removed and is discussed in the Appendix. Since a somewhat large proportion of questions had an observable change, this weakly suggests that the content shared does impact how students think about GAI.  

\subsection{Question 1}

\begin{figure}[h!]
    \centering
    \includegraphics[width=\linewidth]{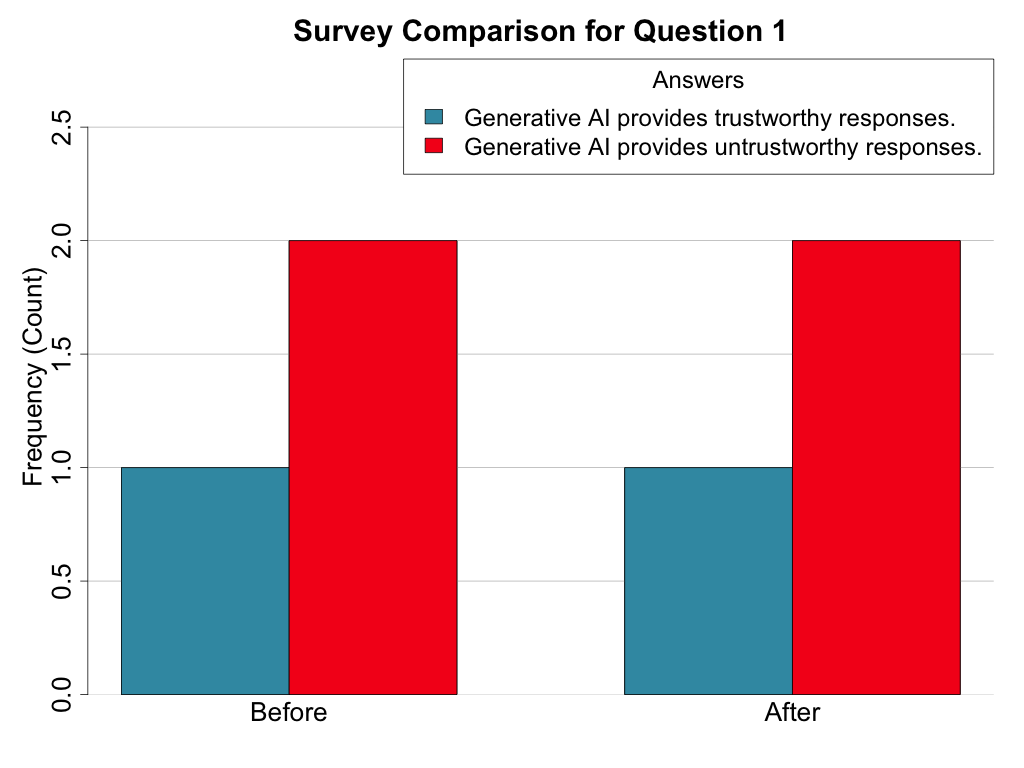}
    \caption{Results for the survey question: Which statement comes closer to your view, even if neither is exactly right? a. Generative AI provides trustworthy responses. b. Generative AI provides untrustworthy responses.}
    \label{fig: q1}
\end{figure}

Figure \ref{fig: q1} provides the results from Question 1.  No students changed their respective minds after watching the lesson.  This suggests that the content shown did not impact the opinions of students regarding trusting GAI.  In fact, most students believe that GAI provides untrustworthy responses.    

\subsection{Question 2}

\begin{figure}[h!]
    \centering
    \includegraphics[width=\linewidth]{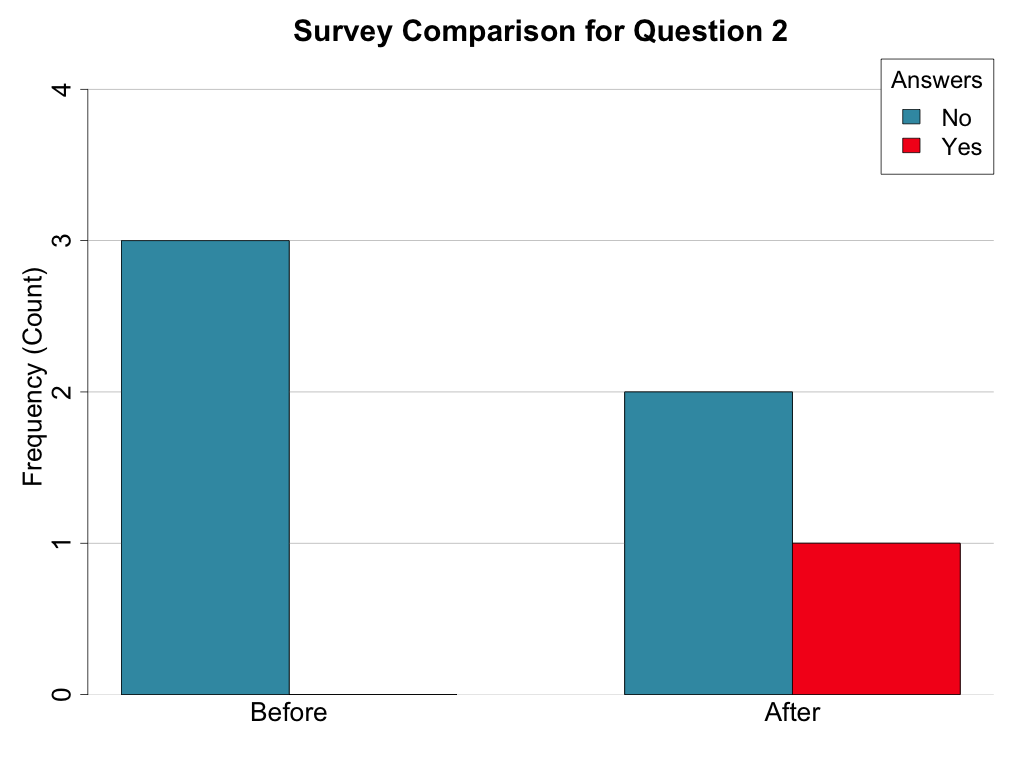}
    \caption{Results for the survey question: Generally speaking, would you say that it is okay to use generative AI to create art in the style of a deceased artist? a. Yes b. No}
    \label{fig: q2}
\end{figure}

Figure \ref{fig: q2} provides the results from Question 2.  This shows that 1 student changed from a `No' to a `Yes' response.  This suggests that the content shown changed the given student's mind to be more accepting of GAI use.   

\subsection{Question 3}

\begin{figure}[h!]
    \centering
    \includegraphics[width=\linewidth]{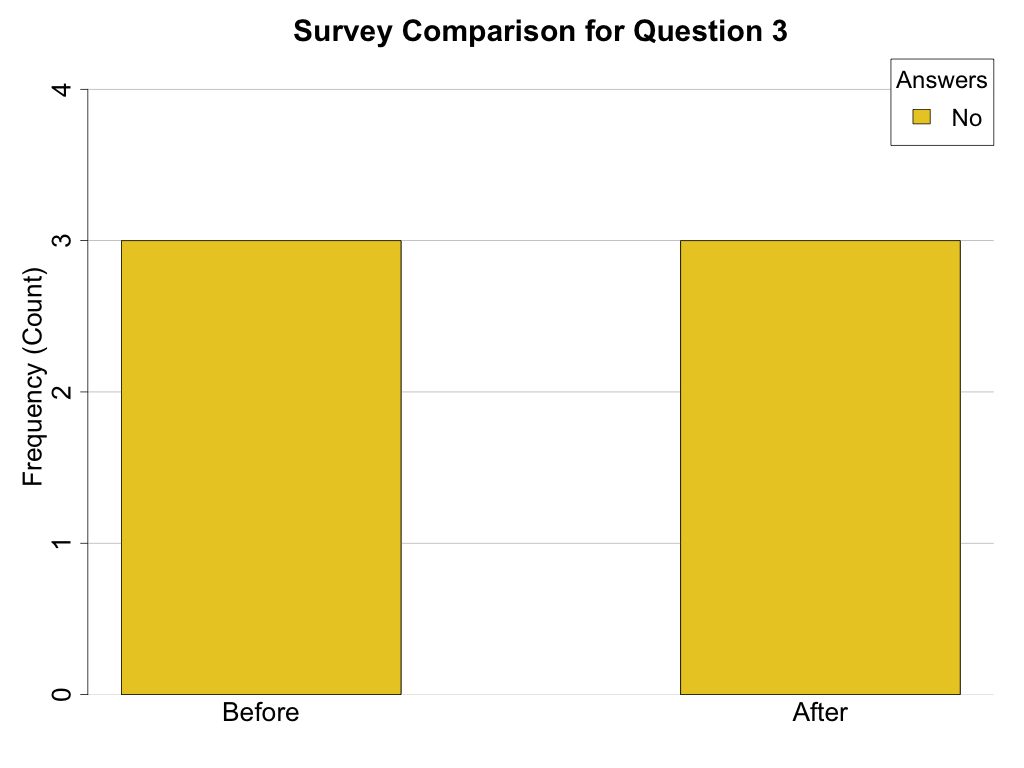}
    \caption{Results for the survey question: Generally speaking, would you say that it is okay to use generative AI to create art in the style of an artist who is alive? a. Yes b. No}
    \label{fig: q3}
\end{figure}

Figure \ref{fig: q3} provides the results from Question 3.  No students changed their respective minds after watching the lesson.  This suggests that the content shown did not impact the opinions of students regarding the art style of creators who are still alive.   

\subsection{Question 4}

Figure \ref{fig: q2} provides the results from Question 2.  This shows that 1 student changed from a `Not Sure' to a `It is acceptable for instructors in my class to use generative AI for any task.' response.  This suggests that the content shown changed the given student's mind regarding instructor use of GAI.  Further, the results indicate that there are is no consencus about instructor use of GAI in the class.    

\begin{figure}[h!]
    \centering
    \includegraphics[width=\linewidth]{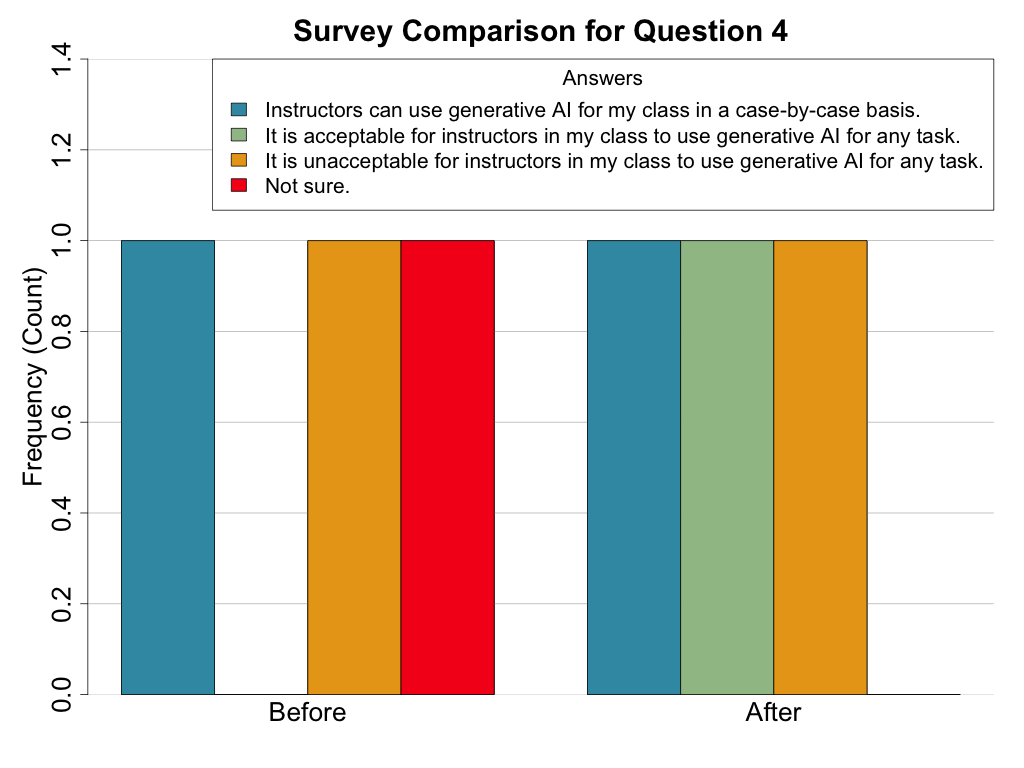}
    \caption{Results for the survey question: Which of the following statements regarding instructors, generative AI, and the classroom use do you agree with the most? a. It is acceptable for instructors in my class to use generative AI for any task. b. It is unacceptable for instructors in my class to use generative AI for any task. c. Instructors can use generative AI for my class in a case-by-case basis. d. Not sure.}
    \label{fig: q4}
\end{figure}

\subsection{Question 6}

\begin{figure}[h!]
    \centering
    \includegraphics[width=\linewidth]{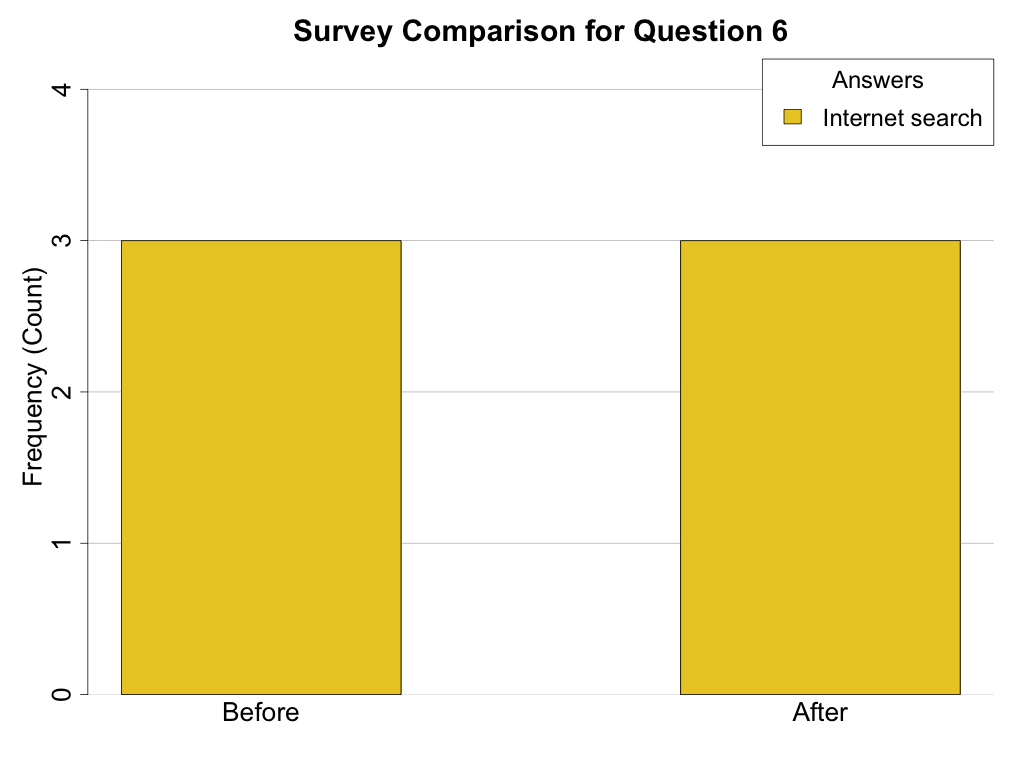}
    \caption{Results for the survey question: Which of the following provides the most factual information? a. Generative AI (i.e., ChatGPT, Gemini, etc.) b. Internet search (i.e., Google, Bing, etc.) c. Neither}
    \label{fig: q6}
\end{figure}

Figure \ref{fig: q6} provides the results from Question 6.  No students changed their respective minds after watching the lesson.  This suggests that the content shown did not impact the opinions of students regarding that an internet search is still perceived to be more factual than a GAI result.  

\subsection{Question 7}

\begin{figure}[h!]
    \centering
    \includegraphics[width=\linewidth]{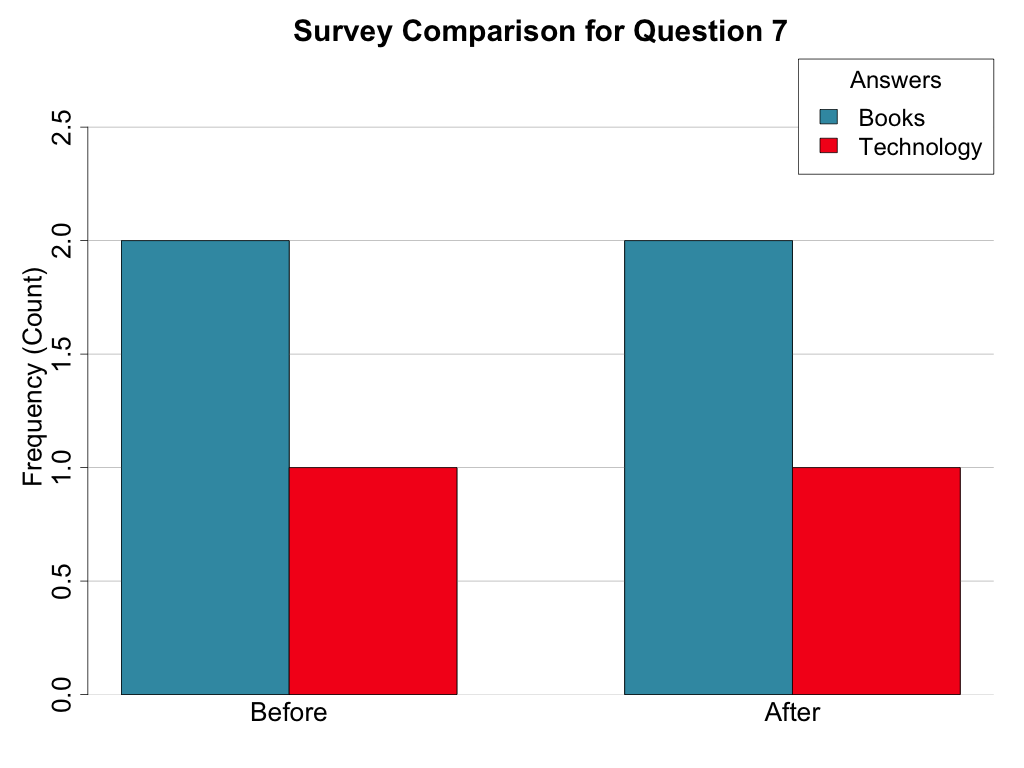}
    \caption{Results for the survey question: What is the most reliable source of factual information? a. People b. Technology c. Books d. Visual media e. Video games}
    \label{fig: q7}
\end{figure}

Figure \ref{fig: q7} provides the results from Question 7.  No students changed their respective minds after watching the lesson.  This suggests that the content shown did not impact the opinions of students regarding that a books and then technology are perceived to be more reliable of factual information.  

\section{Discussion}

\subsection{Response Rate is Lower Than Anticipated}

Past work suggested that we should expect about 9 to 10 students to participate in this survey.  However, we only observed a total of 3 students to allow us to use their data for research.  Our CIs suggest that this is not due to random chance, but a change.  The reason for this is not known, but some educated guesses are proposed herein.  

One is related to when the request for using student data was made.  The Summer 2025 pilot study requested this prior to the final, while the Fall 2025 pilot study requested this after the final. This is the biggest change in the procedure followed from these two studies.  This change occurred in an effort to be extremely conservative and provide students a greater sense that they are not coerced in any manner in participating.  

Another potential reason is that this pilot study, while related to the Summer 2025 study, is a fundamentally different question.  Thus, perhaps this change is the culprit.  The Summer 2025 study is being replicated and analyzed as this paper is being written.  More can be said about this fact in subsequent work. 

Regardless of the reason, a participation rate of about 4\% suggests that classes with larger sections are required to collect a reasonable amount of data to make inferences on the broader student population.  While we were able to pool sections together in this study (since we had the same instructor using the same teaching modality), pooling sections together may prove practically and logistically difficult.  

\subsection{Na\"ive Insights from Survey}

It is difficult to make any definitive inference based on a small sample size.  However, the results do suggest that repeating this experiment could provide important insights on student opinions regarding GAI and how instructor's lessons could impact those opinions.  

\section{Conclusion}   

This study provides valuable insights for future experiments.  With a reported 4.4\% participation rate, this suggests that larger class sizes are need to perform research related to GAI.  A concurrent study from Fall 2025 and planned studies for Spring 2026 semesters will provide further insight on this.  Due to the small sample size, statistically significant insights regarding the survey responses at this time are unwise.  However, this pilot helps to pave the way for greater knowledge on student opinions of GAI to help inform the presentation of GAI topics to students. 

\section{Acknowledgments}

The following tools were used to write this manuscript: Overleaf, Writefull (which is built into Overleaf as of 2024), Microsoft Word, Notability on an iPad, 1st generation Apple Pencil, Macbook Pro personal and work computers, Ubuntu desktop homebuilt desktop computer, Linux Mint desktop home built computer, PatriotAI, Zotero, Preview on Mac, Safari, Vivaldi, and Firefox.  Tools not mentioned were unintentionally omitted.  

\section{Author Contributions}

Dr. Lamberti is the PI and corresponding author of this study. He lead the research by submitting IRB related documents, overseeing student researchers, wrote the paper, edited the paper, taught statistics concepts to student researchers, and led funding efforts. He also ran the CDS 130 courses related to this study.

Samantha Rose Lawrence is a student researcher who conducted the literature review, contributed to writing and editing the manuscript, and collected the data.

Sunbin Kim is a student researcher who contributed to writing and editing the manuscript and code. She also served as the lead teaching assistant for CDS 130.

\clearpage

\bibliographystyle{IEEEtran}
\bibliography{main.bib}

@misc{lamberti_pilot_2025,
	title = {Pilot {Study} on {Generative} {AI} and {Critical} {Thinking} in {Higher} {Education} {Classrooms}},
	copyright = {All rights reserved},
	url = {http://arxiv.org/abs/2509.00167},
	doi = {10.48550/arXiv.2509.00167},
	urldate = {2025-11-12},
	publisher = {arXiv},
	author = {Lamberti, W. F. and Lawrence, S. R. and White, D. and Kim, S. and Abdullah, S.},
	month = sep,
	year = {2025},
	note = {arXiv:2509.00167 [cs]},
}

@article{sax_assessing_2003,
	title = {Assessing {Response} {Rates} and {Nonresponse} {Bias} in {Web} and {Paper} {Surveys}},
	volume = {44},
	doi = {10.1023/A:1024232915870},
	journal = {Research in Higher Education},
	author = {Sax, Linda and Gilmartin, Shannon and Bryant, Alyssa},
	month = aug,
	year = {2003},
	pages = {409--432},
}

@article{wang_learning_2025,
	title = {A learning module for generative {AI} literacy in a biomedical engineering classroom},
	volume = {10},
	issn = {2504-284X},
	url = {https://www.frontiersin.org/journals/education/articles/10.3389/feduc.2025.1551385/full},
	doi = {10.3389/feduc.2025.1551385},
	language = {English},
	urldate = {2026-01-07},
	journal = {Frontiers in Education},
	author = {Wang, Xianglong and Chan, Tiffany Marie and Tamura, Angelika Aldea},
	month = mar,
	year = {2025},
	note = {Publisher: Frontiers},
}

@article{clopper_use_1934,
	title = {The {Use} of {Confidence} or {Fiducial} {Limits} {Illustrated} in the {Case} of the {Binomial}},
	volume = {26},
	issn = {0006-3444},
	url = {https://www.jstor.org/stable/2331986},
	doi = {10.2307/2331986},
	number = {4},
	urldate = {2026-01-05},
	journal = {Biometrika},
	author = {Clopper, C. J. and Pearson, E. S.},
	year = {1934},
	note = {Publisher: [Oxford University Press, Biometrika Trust]},
	pages = {404--413},
}

@article{newcombe_two-sided_1998,
	title = {Two-sided confidence intervals for the single proportion: comparison of seven methods},
	volume = {17},
	copyright = {Copyright © 1998 John Wiley \& Sons, Ltd.},
	issn = {1097-0258},
	shorttitle = {Two-sided confidence intervals for the single proportion},
	url = {https://onlinelibrary.wiley.com/doi/abs/10.1002/%28SICI%291097-0258%2819980430%2917%3A8%3C857%3A%3AAID-SIM777%3E3.0.CO%3B2-E},
	doi = {10.1002/(SICI)1097-0258(19980430)17:8<857::AID-SIM777>3.0.CO;2-E},
	language = {en},
	number = {8},
	urldate = {2026-01-05},
	journal = {Statistics in Medicine},
	author = {Newcombe, Robert G.},
	year = {1998},
	note = {\_eprint: https://onlinelibrary.wiley.com/doi/pdf/10.1002/\%28SICI\%291097-0258\%2819980430\%2917\%3A8\%3C857\%3A\%3AAID-SIM777\%3E3.0.CO\%3B2-E},
	pages = {857--872},
}

@article{thulin_cost_2014,
	title = {The cost of using exact confidence intervals for a binomial proportion},
	volume = {8},
	issn = {1935-7524},
	url = {http://arxiv.org/abs/1303.1288},
	doi = {10.1214/14-EJS909},
	number = {1},
	urldate = {2026-01-05},
	journal = {Electronic Journal of Statistics},
	author = {Thulin, Måns},
	month = jan,
	year = {2014},
	note = {arXiv:1303.1288 [math]},
}

@book{johnson_univariate_2005,
	edition = {1},
	series = {Wiley {Series} in {Probability} and {Statistics}},
	title = {Univariate {Discrete} {Distributions}},
	copyright = {http://doi.wiley.com/10.1002/tdm\_license\_1.1},
	isbn = {978-0-471-27246-5 978-0-471-71581-8},
	url = {https://onlinelibrary.wiley.com/doi/book/10.1002/0471715816},
	language = {en},
	urldate = {2026-01-05},
	publisher = {Wiley},
	author = {Johnson, Norman L. and Kemp, Adrienne W. and Kotz, Samuel},
	month = aug,
	year = {2005},
	doi = {10.1002/0471715816},
}

@article{fosnacht_how_2017,
	title = {How {Important} are {High} {Response} {Rates} for {College} {Surveys}?},
	volume = {40},
	issn = {1090-7009},
	url = {https://muse.jhu.edu/pub/1/article/640611},
	number = {2},
	urldate = {2026-01-05},
	journal = {The Review of Higher Education},
	author = {Fosnacht, Kevin and Sarraf, Shimon and Howe, Elijah and Peck, Leah K.},
	year = {2017},
	note = {Publisher: Johns Hopkins University Press},
	pages = {245--265},
}

@article{wu_response_2022,
	title = {Response rates of online surveys in published research: {A} meta-analysis},
	volume = {7},
	issn = {2451-9588},
	shorttitle = {Response rates of online surveys in published research},
	url = {https://www.sciencedirect.com/science/article/pii/S2451958822000409},
	doi = {10.1016/j.chbr.2022.100206},
	urldate = {2026-01-05},
	journal = {Computers in Human Behavior Reports},
	author = {Wu, Meng-Jia and Zhao, Kelly and Fils-Aime, Francisca},
	month = aug,
	year = {2022},
	pages = {100206},
}

@misc{lamberti_artificial_2024,
	title = {Artificial {Intelligence} {Policy} {Framework} for {Institutions}},
	copyright = {All rights reserved},
	url = {http://arxiv.org/abs/2412.02834},
	doi = {10.48550/arXiv.2412.02834},
	urldate = {2024-12-05},
	publisher = {arXiv},
	author = {Lamberti, William Franz},
	month = dec,
	year = {2024},
	note = {arXiv:2412.02834 [cs]},
}

@misc{jeegandhi2025topic,
      title={When Algorithms Meet Artists: Topic Modeling the AI-Art Debate, 2013-2025}, 
      author={Ariya Mukherjee-Gandhi and Oliver Muellerklein},
      year={2025},
      eprint={2508.03037},
      archivePrefix={arXiv},
      primaryClass={cs.CL},
      url={https://arxiv.org/abs/2508.03037}, 
}

@article{ivanova2025ai,
  title={AI, art and morality},
  author={Ivanova, Milena},
  journal={AI and Ethics},
  pages={1--10},
  year={2025},
  publisher={Springer}
}

@ARTICLE{mahmoud2025,
  author={Mahmoud, Ali B. and Kumar, V and Spyropoulou, Stavroula},
  journal={IEEE Transactions on Engineering Management}, 
  title={Identifying the Public's Beliefs About Generative Artificial Intelligence: A Big Data Approach}, 
  year={2025},
  volume={72},
  number={},
  pages={827-841},
  doi={10.1109/TEM.2025.3534088}}

@misc{fuentes_how_2025,
  title = {How {Americans} {View} {AI} and {Its} {Impact} on {People} and {Society}},
  url = {https://www.pewresearch.org/science/2025/09/17/how-americans-view-ai-and-its-impact-on-people-and-society/},
  language = {en-US},
  urldate = {2025-12-09},
  journal = {Pew Research Center},
  author = {Fuentes, Eileen Yam and Kikuchi, Emma and Pula, Isabelle and Javier, Brian Kennedy},
  month = sep,
  year = {2025},
}

@article{hill_professors_2025,
	chapter = {Technology},
	title = {The {Professors} {Are} {Using} {ChatGPT}, and {Some} {Students} {Aren}’t {Happy} {About} {It}},
	issn = {0362-4331},
	url = {https://www.nytimes.com/2025/05/14/technology/chatgpt-college-professors.html},
	language = {en-US},
	urldate = {2025-11-05},
	journal = {The New York Times},
	author = {Hill, Kashmir},
	month = may,
	year = {2025},
}

@article{kircher_people_2025,
	chapter = {Style},
	title = {People {Love} {Studio} {Ghibli}. {But} {Should} {They} {Be} {Able} to {Recreate} {It}?},
	issn = {0362-4331},
	url = {https://www.nytimes.com/2025/03/27/style/ai-chatgpt-studio-ghibli.html},
	language = {en-US},
	urldate = {2025-12-08},
	journal = {The New York Times},
	author = {Kircher, Madison Malone},
	month = mar,
	year = {2025},
}

@article{Hassine_Neeman_2019, title={The Zombification of art history: how AI resurrects dead masters, and perpetuates historical biases}, volume={11}, url={https://revistas.ucp.pt/index.php/jsta/article/view/7331}, DOI={10.7559/citarj.v11i2.663}, abstractNote={&amp;lt;p&amp;gt;In the past few years deep-learning AI neural networks have achieved major milestones in artistic image analysis and generation, producing what some refer to as ‘art.’ We reflect critically on some of the artistic shortcomings of a few projects that occupied the spotlight in recent years. We introduce the term ‘Zombie Art’ to describe the generation of new images of dead masters, as well as ‘The AI Reproducibility Test.’ We designate the problems inherent in AI and in its application to art history. In conclusion, we propose new directions for both AI-generated art and art history, in the light of these new powerful AI technologies of artistic image analysis and generation.&amp;lt;/p&amp;gt;}, number={2}, journal={Journal of Science and Technology of the Arts}, author={Hassine, Tsila and Neeman, Ziv}, year={2019}, month={May}, pages={28–35} }

@Online{cengel_2024_smithsonian,
 author = {Cengel, Katya},
 year = {2024},
 title = {The First A.I.-Generated Art Dates Back to the 1970s},
 journal = {Smithsonian Magazine},
 url = {https://www.smithsonianmag.com/innovation/first-ai-generated-art-dates-back-to-1970s-180983700/},
 urldate = {2025-12-11}
}

@Online{nytroose2022,
 author = {Roose, Kevin},
 year = {2022},
 title = {An A.I.-Generated Picture Won an Art Prize. Artists Aren’t Happy.},
 journal = {The New Work Times},
 url = {https://www.nytimes.com/2022/09/02/technology/ai-artificial-intelligence-artists.html},
 urldate = {2025-12-11}
}

@article{grynbaum_mac_2023aicopyright,
  author  = "Grynbaum, Mac",
  title   = "The Times Sues OpenAI and Microsoft Over A.I. Use of Copyrighted Work",
  journal = "The New York Times",
  year    = 2023
}

@article{barnes2025disney,
  author  = "Barnes",
  title   = "Disney and Universal Sue A.I. Firm for Copyright Infringement",
  journal = "The New York Times",
  year    = 2025
}

@misc{thierer_artificial_2017,
	address = {Rochester, NY},
	type = {{SSRN} {Scholarly} {Paper}},
	title = {Artificial {Intelligence} and {Public} {Policy}},
	url = {https://papers.ssrn.com/abstract=3021135},
	doi = {10.2139/ssrn.3021135},
	language = {en},
	urldate = {2024-10-30},
	publisher = {Social Science Research Network},
	author = {Thierer, Adam D. and Castillo O'Sullivan, Andrea and Russell, Raymond},
	month = aug,
	year = {2017},
}

@misc{midjourney2024guide,
	title = {Getting Started Guide}}

@article{goodfellow_generative_2014,
	title = {Generative {Adversarial} {Networks}},
	url = {http://arxiv.org/abs/1406.2661},
	language = {en},
	urldate = {2018-12-31},
	journal = {arXiv:1406.2661 [cs, stat]},
	author = {Goodfellow, Ian J. and Pouget-Abadie, Jean and Mirza, Mehdi and Xu, Bing and Warde-Farley, David and Ozair, Sherjil and Courville, Aaron and Bengio, Yoshua},
	month = jun,
	year = {2014},
	note = {arXiv: 1406.2661},
}

@book{bhattacharyya_statistical_1977,
	edition = {1st},
	title = {Statistical {Concepts} and {Methods}},
	url = {https://www.wiley.com/en-us/Statistical+Concepts+and+Methods-p-9780471072041},
	language = {en-us},
	urldate = {2018-07-20},
	publisher = {Wiley},
	author = {Bhattacharyya, Gouri K. and Johnson, Richard A.},
	year = {1977},
}

\appendix

\section*{IRB Details}

The research plan for the institutional review board (IRB) was conducted at GMU via the Office of Research Integrity and Assurance.  The review and feedback was provided via Research Administration Management Portal (RAMP).  Note that requests for permission to use student data were requested after the final was taken.  

\section*{Details on Treatment}

Students were given a multiple choice assignment consisting of several multiple choice questions.  In the middle of the assignment, the students were requested to watch a video.  They were then asked the same questions after they watched the video. Differences in the opinions were noted after watching the video. 

\section*{GitHub}

Additional resources such as the code and data used for this study are available on our GitHub.  The link is \url{https://github.com/billyl320/gai_student_opinion}.

\section*{Questions Asked of Students}

Below are the questions asked to the students in the participating sections of CDS 130. 

\begin{enumerate}
    \item Which statement comes closer to your view, even if neither is exactly right?
    \begin{enumerate}
        \item Generative AI provides trustworthy responses.
        \item Generative AI provides untrustworthy responses. 
    \end{enumerate}
    \item Generally speaking, would you say that it is okay to use generative AI to create art in the style of a deceased artist?
    \begin{enumerate}
        \item Yes
        \item No
    \end{enumerate}
    \item Generally speaking, would you say that it is okay to use generative AI to create art in the style of an artist who is alive?
    \begin{enumerate}
        \item Yes
        \item No
    \end{enumerate}
    \item Which of the following statements regarding instructors, generative AI, and the classroom use do you agree with the most?
    \begin{enumerate}
        \item It is acceptable for instructors in my class to use generative AI for any task.
        \item It is unacceptable for instructors in my class to use generative AI for any task.
        \item Instructors can use generative AI for my class in a case-by-case basis.
        \item Not sure.
    \end{enumerate}
    \item Which of the following statements regarding students, generative AI, and the classroom use do you agree with the most?
    \begin{enumerate}
        \item It is acceptable for students in my class to use generative AI for any task.
        \item It is unacceptable for students in my class to use generative AI for any task.
        \item Students should only be able to use generative AI for my class in a case-by-case basis.
        \item Not sure.
    \end{enumerate}
    \item Which of the following provides the most factual information?
    \begin{enumerate}
        \item Generative AI (i.e., ChatGPT, Gemini, etc.)
        \item Internet search (i.e., Google, Bing, etc.)
        \item Neither
    \end{enumerate}
    \item What is the most reliable source of factual information?
    \begin{enumerate}
        \item People
        \item Technology
        \item Books
        \item Visual media
        \item Video games
    \end{enumerate}
\end{enumerate}

\subsection{Question 5}

Figure \ref{fig: q5} provides the results from Question 5.  This question was discarded due to an error in the creation of the survey as seen in Figure \ref{fig: q5_canvas}.  The question was intended to be about students and GAI, but some of the responses were related to instructors.  Thus, we removed this question from the pilot study analysis to be conservative.

\begin{figure}[h!]
    \centering
    \includegraphics[width=\linewidth]{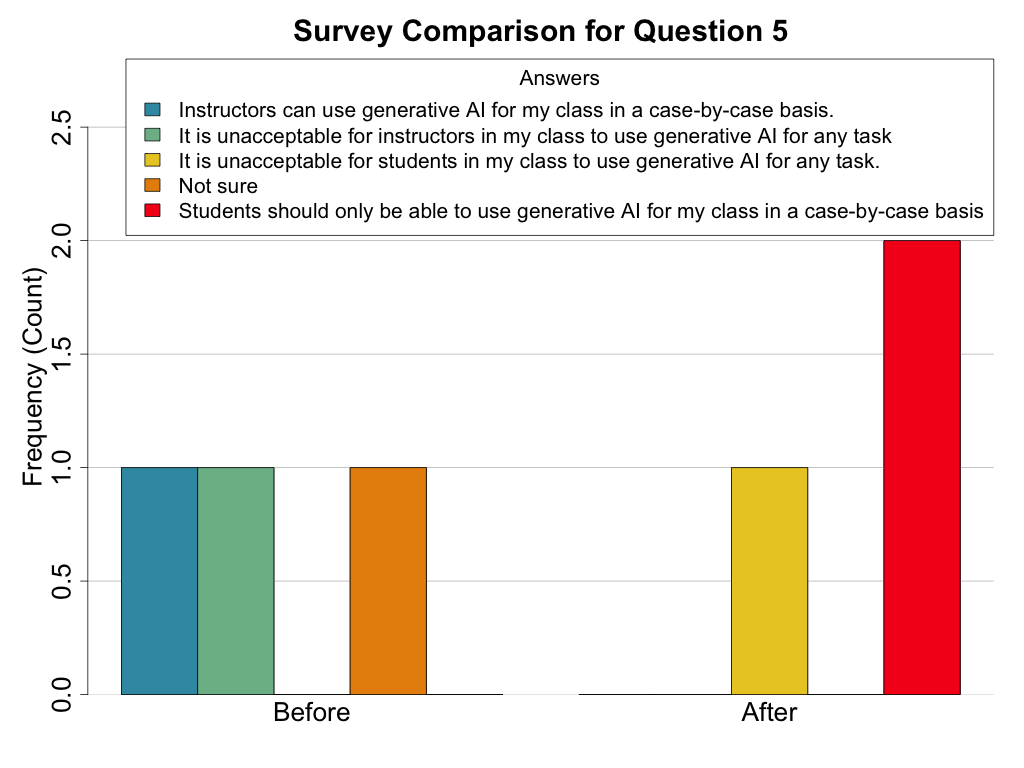}
    \caption{Results for the survey question: Which of the following statements regarding students, generative AI, and the classroom use do you agree with the most? a. It is acceptable for students in my class to use generative AI for any task. b. It is unacceptable for students in my class to use generative AI for any task. c. Students should only be able to use generative AI for my class in a case-by-case basis. d. Not sure.}
    \label{fig: q5}
\end{figure}

\begin{figure}[h!]
    \centering
    \includegraphics[width=\linewidth]{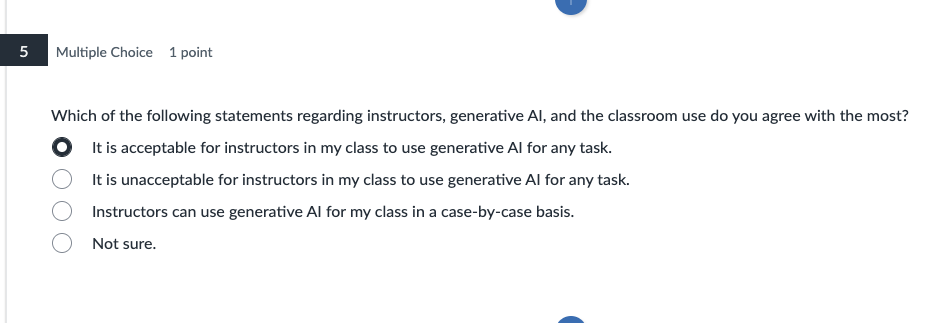}
    \caption{Question 5 in Canvas where the wording was incorrect and confusing.}
    \label{fig: q5_canvas}
\end{figure}

\end{document}